# Causal Discovery from Changes

Jin Tian and Judea Pearl
Cognitive Systems Laboratory
Computer Science Department
University of California, Los Angeles, CA 90024
{*jtian, judea* }@cs.ucla.edu

## Abstract

We propose a new method of discovering causal structures, based on the detection of local, spontaneous changes in the underlying data-generating model. We analyze the classes of structures that are equivalent relative to a stream of distributions produced by local changes, and devise algorithms that output graphical representations of these equivalence classes. We present experimental results, using simulated data, and examine the errors associated with detection of changes and recovery of structures.

## 1 Introduction

In recent years, several graph-based algorithms have been developed for the purpose of inferring causal structures from empirical data. Some are based on detecting patterns of conditional independence relationships [Pearl and Verma, 1991, Spirtes et al., 1993], and some are based on Bayesian approaches [Cooper and Herskovits, 1992, Heckerman et al., 1995]. These discovery methods assume static environment, that is, a time-invariant distribution and a time-invariant data-generating model, and attempt to infer structures that encode dynamic aspects of the environment, for example, how probabilities would change as a result of interventions. This transition, from static to dynamic information, constitutes a major inferential leap, and is severely limited by the inherent indistinguishability (or equivalence) relation that governs Bayesian networks [Verma and Pearl, 1990].

One way of overcoming this basic limitation is to augment the data with partial causal knowledge, if such is available. [Spirtes et al., 1993], for example, discussed the use of experimental data to identify causal relationships. [Cooper and Yoo, 1999] discussed a Bayesian method of causal discovery from a mixture of observational and experimental data.

We propose a new method of discovering causal relations in data, based on the detection and interpretation of local spontaneous changes in the environment. While previous methods assume that data are generated by a static statistical distribution, our proposal aims at exploiting dynamic changes in that distribution. Such changes are always present in any realistic domain that is embedded in a larger background of dynamically changing conditions. For example, natural disasters, armed conflicts, epidemics, labor disputes, and even mundane decisions by other agents, are unexpected eventualities that are not naturally captured in distribution functions. The occurrence of such eventualities tend to *alter* the distribution under study and yield changes that are markedly different from ordinary statistical fluctuations. Whereas static analysis views these changes as nuisance, and attempts to adjust and compensate for them, we will view them as a valuable source of information about the data-generating process. A controlled experimental study may be thought of as a special case of these environmental changes, where the external influence involves fixing a designated variable to some predetermined value. In general, however, the external influence may be milder, merely changing the conditional probability of a variable, given its causes. Moreover, in marked contrast to controlled experiments, we may not know in advance the nature of the change, its location, or even whether it took place; these may need to be inferred from the data itself.

The basic idea has its roots in the economic literature. The economist Kevin Hoover (1990) attempted to infer the direction of causal influences among economic variables (e.g., employment and money supply) by observing the changes that sudden modifications in the economy (e.g., tax reform, labor dispute) induced in the statistics of these variables. Hoover assumed that the conditional probability of an effect given its causes remains invariant to changes in the mechanism that generates the cause, while the conditional probability of a cause given the effect would not remain invariant under such changes. This asymmetry may be useful in distinguishing cause and effect.

Today we understand more precisely the conditions under which such asymmetries would prevail and how to interpret such asymmetries in the context of large, multi-variate systems. Whenever we obtain reliable information (e.g., from historical or institutional knowledge) that an abrupt local change has taken place in a specific mechanism that constrains a given family of variables, we can use the observed changes in the marginal and conditional probabilities surrounding those variables to determine the direction of causal influences in the domain. The statistical features that remain invariant under such changes, as well as the causal assumptions underlying this invariance, are encoded in the causal diagram at hand, and can be used therefore for testing the validity of a given structure. Likewise, conflicts between observed and predicted changes can be used for automatic restructuring of the topology of the structure at hand.

## 2 Causal Models and Mechanism Change

Let our problem domain be a set of discrete random variables $V = \{V_1, \ldots, V_n\}$. We assume that a *causal model* over $V$ is a pair $M = <G, \Theta_G>$, where $G$ is a DAG over $V$, called a *causal diagram*, and $\Theta_G$ is a set of probability parameters. We assume that each variable $V_i$ can take values from a finite domain, $Dm(V_i) = \{v_{i1}, \ldots, v_{ir_i}\}$, where $r_i$ is the number of states of $V_i$. We use $Pa_i$ to represent the set of parents of $V_i$ in a causal diagram $G$ and $Dm(Pa_i)$ to represent the set of states of $Pa_i$. Let $\theta_{v_i;pa_i}, v_i \in Dm(V_i), pa_i \in Dm(Pa_i)$ denote the multinomial parameter corresponding to the conditional probability $P(v_i|pa_i)$. We will use the following notations: $\vec{\theta}_{pa_i} = \{\theta_{v_i;pa_i}|v_i \in Dm(V_i)\}$, $\Psi_i = \cup_{pa_i \in Dm(Pa_i)} \vec{\theta}_{pa_i}$, $\Theta_G = \cup_{i=1}^n \Psi_i$. Assuming the Causal Markov condition [Spirtes et al., 1993], a causal model $M = <G, \Theta_G>$ generates a probability distribution

$$P(v) = \prod_i \theta_{v_i;pa_i}. \qquad (1)$$

A probability distribution $P(V)$ is said to be *compatible* with a causal diagram $G$ if $P(V)$ can be generated by some causal model $M = <G, \Theta_G>$.

The factorization in Eq. (1) obtains causal character through the assumption of *modularity*; each family in the causal diagram represents an autonomous physical mechanism and is subjected to change without influencing other mechanisms. We formally define mechanism change as follows.

**Definition 1 (Mechanism Change)** *A mechanism change to a causal model $M = <G, \Theta_G>$ at a variable $V_i$ is a transformation of $M$ that produces a new model, $M_{V_i} = <G, \Theta'_G>$, where $\Theta'_G = \Psi'_i \cup (\Theta_G \setminus \Psi_i)$ and $\Psi'_i$ is a set of parameters having values that differ from those in $\Psi_i$.*

We assume in this paper that the parent set $Pa_i$ does not change in a mechanism change. An intervention that fixes $V_i$ to a particular value is a special case of a mechanism change. Let $P(V)$ be the distribution generated by $M$, as in Eq. (1). Then the distribution generated by $M_{V_i}$ is given by

$$P_{V_i}(v) = \theta'_{v_i;pa_i} \prod_{j \neq i} \theta_{v_j;pa_j}. \qquad (2)$$

We will call $(P, P_{V_i})$ a *transition pair (TP)* and $V_i$ the *focal variable* of the transition. Assume that a series of mechanism changes occurred successively to a causal model $M = <G, \Theta_G^0>$, and let $F = (V_{i_1}, \ldots, V_{i_k})$ denote the corresponding sequence of focal variables. We use $P_{TS} = (P^0, P^1, \ldots, P^k)$ to denote the sequence of distributions generated by such a series, and call the pair $(P_{TS}, F)$ a *transition sequence (TS)*.

As oracles for cause-and-effect relations, causal models can predict the effects that any external or spontaneous changes have on the distributions. Conversely, by detecting how probability distributions change under various mechanism changes, we obtain information on the structure of the model generating those distributions. In this paper, we assume that we are given a TS $(P_{TS}, F)$ corresponding to some causal diagram $G$, and our task is to recover $G$. We will then assume that we have a sequence of datasets $\mathbb{D}_{TS} = \{D^0, \ldots, D^k\}$, where each $D^i$ is a set of random samples from a distribution $P^i$, such that each pair $(P^{j-1}, P^j)$ is a TP with focal variable $V_{i_j}$, and our task will be to recover a causal diagram (or a set of diagrams) that can generate $\mathbb{D}_{TS}$. First, we study what can be learned from a TS.

## 3 Indistinguishability of Causal Diagrams

In this section, we study the classes of causal structures that are indistinguishable (or "equivalent") relative to a TS.

The statistical information provided by any causal diagram is completely encoded in the independence relationships among the variables. Therefore, two causal diagrams are statistically indistinguishable given one static distribution if and only if they are independence equivalent. The graphical conditions for independence equivalence are given by the following theorem.

**Theorem 1 (Independence Equivalence)** *Two causal diagrams are independence equivalent if and only if they have the same skeletons and the same sets of v-structures, that is, two converging arrows whose tails are not connected by an arrow* (Verma and Pearl 1990).

Now assume that we have a TP with focal variable $V_i$. A causal diagram $G$ is said to be *compatible with a transition pair* $(P, P_{V_i})$ if $P$ can be generated by a causal model $M = <G, \Theta_G>$ and $P_{V_i}$ can be generated by a causal model $M_{V_i} = <G, \Theta'_G>$ resulted from a mechanism change to $M$ at $V_i$. Note that a causal diagram could be compatible with both $P$ and $P_{V_i}$ but *not* compatible with the TP $(P, P_{V_i})$. Among those independence-equivalent diagrams compatible with both $P$ and $P_{V_i}$, a TP $(P, P_{V_i})$ can distinguish those that can generate $P_{V_i}$ from $P$ with a *single* mechanism change from those that can not. Two causal diagrams $G_1$ and $G_2$ are called *transition pair equivalent* with respect to a TP with focal variable $V_i$, or $V_i$-*transition equivalent*, if every TP $(P, P_{V_i})$ compatible with $G_1$ is also compatible with $G_2$. Two causal diagrams are statistically indistinguishable given a TP $(P, P_{V_i})$ if and only if they are $V_i$-transition equivalent.

**Theorem 2 (Transition Pair Equivalence)** *Two causal diagrams $G_1$ and $G_2$ are $V_i$-transition equivalent if and only if they have the same skeletons, the same sets of v-structures, and the same sets of parents for $V_i$.*

*Proof:* Let $G_1$ be compatible with a TP $(P, P_{V_i})$. $G_2$ must have the same skeletons and the same sets of v-structures as $G_1$ to be compatible with $P$ (and $P_{V_i}$) by Theorem 1. We have the following decomposition:

$$P(v) = P(v_i|pa_i^1)\prod_{j\neq i} P(v_j|pa_j^1) = P(v_i|pa_i^2)\prod_{j\neq i} P(v_j|pa_j^2), \quad (3)$$

where $Pa_i^1$ and $Pa_i^2$ are parents of $V_i$ in $G_1$ and $G_2$ respectively. $G_1$ is compatible with the TP $(P, P_{V_i})$, hence can generate $P_{V_i}$ from $P$ by a mechanism change at $V_i$:

$$P_{V_i}(v) = P_{V_i}(v_i|pa_i^1)\prod_{j\neq i} P(v_j|pa_j^1). \quad (4)$$

Plugging the expression for $\prod_{j\neq i} P(v_j|pa_j^1)$ from Eq. (3) into Eq. (4), we have

$$P_{V_i}(v) = P_{V_i}(v_i|pa_i^1)\frac{P(v_i|pa_i^2)}{P(v_i|pa_i^1)}\prod_{j\neq i} P(v_j|pa_j^2). \quad (5)$$

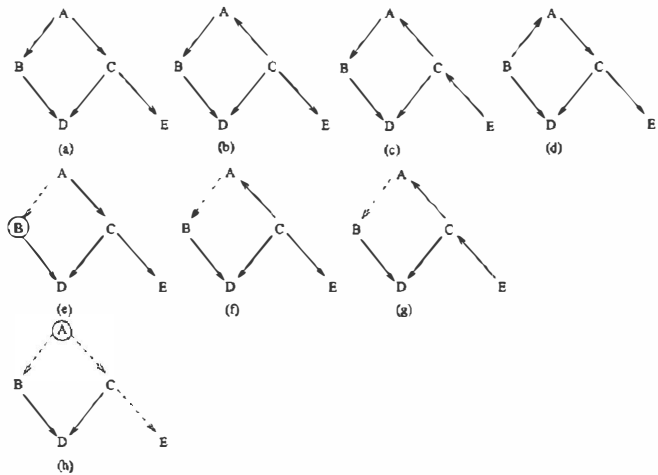

Figure 1: TS equivalence. Assuming (a) is the actual causal diagram. (a)-(d) are independence equivalent. (e)-(g) are $B$-transition equivalent. A mechanism change on $A$ determines a unique causal diagram (h).

$G_2$ is also compatible with the transition pair $(P, P_{V_i})$ if and only if

$$P_{V_i}(v) = P_{V_i}(v_i|pa_i^2)\prod_{j\neq i} P(v_j|pa_j^2). \quad (6)$$

Eqs. (5) and (6) lead to

$$P_{V_i}(v_i|pa_i^1)\frac{P(v_i|pa_i^2)}{P(v_i|pa_i^1)} = P_{V_i}(v_i|pa_i^2), \quad (7)$$

which holds for any distribution $P$ and $P_{V_i}$ if and only if $G_1$ has the same parent set for $V_i$ as $G_2$ ($Pa_i^1 = Pa_i^2$); if $G_1$ has a different parent set for $V_i$ with $G_2$, Eq. (7) will impose some constraints between $P$ and $P_{V_i}$, and will not hold for arbitrary possible transition pair $(P, P_{V_i})$. □

A TS is simply a series of TP's. Accordingly, we say that a causal diagram is *compatible with a transition sequence* $P_{TS} = (P^0, P^1, \ldots, P^k)$, $F = (V_{i_1}, \ldots, V_{i_k})$ if it is compatible with each TP $(P^{j-1}, P^j)$ in the sequence. Likewise, two causal diagrams $G_1$ and $G_2$ are called *transition sequence equivalent* with respect to a TS $(P_{TS}, F)$, or $F$-*transition equivalent*, if every TS $(P_{TS}, F)$ compatible with $G_1$ is also compatible with $G_2$. Two causal diagrams are statistically indistinguishable given a TS $(P_{TS}, F)$ if and only if they are $F$-transition equivalent.

**Theorem 3 (Transition Sequence Equivalence)** *Two causal diagrams are $F$-transition equivalent if and only if they have the same skeletons, the same sets of v-structures, and the same sets of parents for variables in $F$.*

Theorem 3 says that a TS determines the directions of the edges between the focal variables and their neighbors (among the set of independence-equivalent diagrams). Figure 1 shows an example of TS equivalence. Given a TS, the most we can expect to recover is a set of causal diagrams that are TS-equivalent, as defined by Theorem 3. We may find this equivalence class by detecting independence relations and distribution changes.

## 4 Learning Causation by Detecting Changes

In this section, we identify the causal information that can be learned by detecting various changes in the probability distributions, in particular, changes in the marginal probability of each variable. The following theorem is obvious.

**Theorem 4** *A mechanism change at a variable $X$ to a causal model $M = <G, \Theta_G>$ may alter the marginal probabilities of the descendants of $X$ in $G$ and can not alter the marginals of nondescendants of $X$.*

It is possible of course that, for some peculiar parameter changes, the marginal probabilities of some descendants of $X$ would not change. When recovering causal information from distributional changes, we assume a restriction on a TS called *influentiality*.

**Definition 2 (influentiality)** *A TP $(P, P_X)$ generated by a causal model $<G, \Theta_G>$ is said to be influential if for every descendant $Y$ of $X$ in $G$, the marginal distribution $P_X(Y)$ is different from $P(Y)$. A TS is influential if every TP in the sequence is influential.*

Given a TP $(P, P_X)$, and assuming that we can test each variable for marginal distribution change, we can draw the following inferences. If the marginal of a variable $Y$ has changed, we conclude that $Y$ is a descendant of $X$. If the marginal of a variable $Z$ has not changed, we conclude that $Z$ is a nondescendant of $X$. We thus conclude that $Z < X < Y$ should be a causal order consistent with the causal diagram. Next we discuss how to piece together ordering information of this kind, as obtained from a TS.

### 4.1 Partitioning the variables

Given a TS $P_{TS}$, $F = (V_{i_1}, \ldots, V_{i_k})$, each variable can be characterized by a sequence of 1's and 0's, a tag $a_1, \ldots, a_k$, where $a_i$ reflects whether the marginal of that variable changed ($a_i = 1$) or not ($a_i = 0$) in the $i$th transition of the sequence. Non-focal variables that are given the same tags cannot be distinguished by the TS (through detecting marginal changes), and no information can therefore be extracted about their relative causal order in the causal diagram. We may put all such variables into a bucket labeled with the same tag, denoted by $B_{a_1 \cdots a_k}$. Clearly, since we have no information on causal relations among variables within the same bucket, all variables in a bucket stand in the same ordering relation to all variables in another bucket. Focal variables need special treatment since they carry more information, and we will put each focal variable into an individual bucket called a *focal bucket*, denoted by $B^f_{a_1 \cdots a_k}$.

We classify variables into buckets with the following algorithm.

**Algorithm 1 (Partitioning Variable)**
*Input: a TS $P_{TS}$, $F = (V_{i_1}, \ldots, V_{i_k})$.*
*Output: A set of buckets, each associated with a tag $a_1 \ldots a_k$, and each containing a set of variables.*

*Put all variables in a bucket $B$.*
*For the $i$th mechanism change, $i = 1, \ldots, k$,*
　*For each bucket $B_{a_1 \cdots a_{i-1}}$ including focal buckets*
　　*if it contains the $i$th focal variable, put it in a focal bucket $B^f_{a_1 \cdots a_{i-1} 1}$.*
　　*put other changing variables in $B_{a_1 \cdots a_{i-1} 1}$.*
　　*put non-changing variables in $B_{a_1 \cdots a_{i-1} 0}$.*

We show the partitioning process by an example. Assume that the actual causal diagram is the DAG shown in Figure 2(a) and that we are given a TS $(P, P_X, P_Y)$. In the first transition, with $X$ as the focal variable, $P(Y)$ does not change, hence $B_0 = \{Y\}$; $P(X), P(Z), P(W), P(Q)$ do change, hence we form $B_1 = \{Z, W, Q\}$, $B^f_1 = \{X\}$. Note that a focal variable is put into an individual bucket. In the second transition, with $Y$ as the focal variable, $P(Y)$ changes, giving $B^f_{01} = \{Y\}$; $P(Z)$ and $P(W)$ change, giving $B_{11} = \{Z, W\}$; $P(Q)$ and $P(X)$ do not change, giving $B_{10} = \{Q\}$ and $B^f_{10} = \{X\}$. As a result, the variables are partitioned into four buckets: $B^f_{10} = \{X\}, B^f_{01} = \{Y\}, B_{10} = \{Q\}, B_{11} = \{Z, W\}$.

### 4.2 Extracting causal information

We shall now discuss what causal information we can extract from the tags attached to buckets. Consider any two buckets $B_{a_1 \cdots a_k}$ and $B_{b_1 \cdots b_k}$. If there exists a bit such as $a_i < b_i$ (i.e., $a_i = 0$ and $b_i = 1$), it must be that, in the $i$th transition, the marginals of variables in $B_{a_1 \cdots a_k}$ did not change and the marginals of variables in $B_{b_1 \cdots b_k}$ did. Therefore, no variable in $B_{a_1 \cdots a_k}$ is a descendant of any variable in $B_{b_1 \cdots b_k}$. On the other hand, if there exists another bit such that $a_j > b_j$ ($a_j = 1, b_j = 0$), then no variable in $B_{b_1 \cdots b_k}$ is a descendant of any variable in $B_{a_1 \cdots a_k}$, which means that there exists no directed path, in particular no edge, between any variable in $B_{a_1 \cdots a_k}$ and any vari-able in $B_{b_1\cdots b_k}$. The equality $a_i = b_i, i = 1,\ldots,k$ can only happen if one of the buckets is a focal bucket, in which case the focal variable is an ancestor of all the variables in the other bucket. In summary, the relation between two buckets $B_{a_1\cdots a_k}$ and $B_{b_1\cdots b_k}$ is determined as follows:

R1 $a_i \leq b_i, i = 1,\ldots,k$ and $\exists j, a_j < b_j$: variables in $B_{a_1\cdots a_k}$ are nondescendants of variables in $B_{b_1\cdots b_k}$, denoted by $B_{a_1\cdots a_k} < B_{b_1\cdots b_k}$.

R2 $a_i \geq b_i, i = 1,\ldots,k$ and $\exists j, a_j > b_j$: $B_{b_1\cdots b_k} < B_{a_1\cdots a_k}$.

R3 There exist two bits $i \neq j$ such that $a_i < b_i$ and $a_j > b_j$: there can be no directed path between any variable in $B_{a_1\cdots a_k}$ and any variable in $B_{b_1\cdots b_k}$.

R4 $a_i = b_i, i = 1,\ldots,k$, one of the buckets, say $B^f_{a_1\cdots a_k}$, is a focal bucket: all variables in $B_{b_1\cdots b_k}$ must be descendants of the focal variable in $B^f_{a_1\cdots a_k}$, which is a stronger relation than that in R1 and R2 but will still be denoted by $B^f_{a_1\cdots a_k} < B_{b_1\cdots b_k}$.

The focal buckets convey more information. Let $B_{a_1\cdots a_k}$ be a focal bucket containing the focal variable $V_{i_j}$ for the $j$th transition. Then if $b_j = 1$, we have that all variables in $B_{b_1\cdots b_k}$ are descendants of $V_{i_j}$ since their marginals changed in the $j$th transition. This rule is consistent with the above rules R1–R3, hence it is applied only in R4 when R1–R3 cannot determine a relation. However, in practice, due to imperfect statistical tests, there may be conflicts between them. For example, we may determine that there is no edge between $B_{a_1\cdots a_k}$ and $B_{b_1\cdots b_k}$ by R3 and in the same time $B_{a_1\cdots a_k}$ is a focal bucket for the $j$th transition and $b_j = 1$. These conflicts signal mistakes in the statistical tests, and whenever there are conflicts, we will declare the relation as "unknown". We summarize the above discussions with the following algorithm.

### Algorithm 2 (Extracting Relation)
*Input: two buckets $B_{a_1\cdots a_k}$ and $B_{b_1\cdots b_k}$.*
*Output: the relation between the two buckets, could be "<", "no-directed-path (NDP)", or "unknown".*

1. $a_i \leq b_i, i = 1,\ldots,k$ and $\exists j, a_j < b_j$: if $B_{b_1\cdots b_k}$ is a focal bucket for the $l$th transition and $a_l = 1$ then "unknown", else $B_{a_1\cdots a_k} < B_{b_1\cdots b_k}$.

2. $a_i \geq b_i, i = 1,\ldots,k$ and $\exists j, a_j > b_j$: if $B_{a_1\cdots a_k}$ is a focal bucket for the $l$th transition and $b_l = 1$ then "unknown", else $B_{b_1\cdots b_k} < B_{a_1\cdots a_k}$.

3. There exist two bits $i \neq j$ such that $a_i < b_i$ and $a_j > b_j$: if $B_{b_1\cdots b_k}$ is a focal bucket for the $l$th transition and $a_l = 1$ or $B_{a_1\cdots a_k}$ is a focal bucket for the $l$th transition and $b_l = 1$ then "unknown", else "NDP".

4. $a_i = b_i, i = 1,\ldots,k$: if both buckets are focal buckets then "unknown", else let the focal bucket be $B^f_{a_1\cdots a_k}$, then $B^f_{a_1\cdots a_k} < B_{b_1\cdots b_k}$.

Consider the binary relation "<" on the set of buckets as defined in the Algorithm 2. We have the following theorem.

**Theorem 5** *The binary relation "<" on the set of buckets is a partial order.*

*Proof:* The relation is transitive. If $B_{a_1\cdots a_k} < B_{b_1\cdots b_k}$ and $B_{b_1\cdots b_k} < B_{c_1\cdots c_k}$, we have $a_i \leq b_i \leq c_i, i = 1,\ldots,k$.

1. $\exists j, a_j < c_j$. If $B_{c_1\cdots c_k}$ is not a focal bucket, then we have $B_{a_1\cdots a_k} < B_{c_1\cdots c_k}$. If $B_{c_1\cdots c_k}$ is a focal bucket for the $l$th transition and $a_l = 1$, then $b_l = 1$ since $a_l \leq b_l \leq c_l$, which contradicts $B_{b_1\cdots b_k} < B_{c_1\cdots c_k}$.

2. $a_i = c_i, i = 1,\ldots,k$. Then $a_i = b_i = c_i, i = 1,\ldots,k$, and then $B_{a_1\cdots a_k}$ has to be a focal bucket and $B_{b_1\cdots b_k}$ is not one in order to have the relation $B_{a_1\cdots a_k} < B_{b_1\cdots b_k}$, which then contradicts $B_{b_1\cdots b_k} < B_{c_1\cdots c_k}$.

The relation is antisymmetric. If $B_{a_1\cdots a_k} < B_{b_1\cdots b_k}$ and $B_{b_1\cdots b_k} < B_{a_1\cdots a_k}$, then $a_i = b_i, i = 1,\ldots,k$. Since they cannot both be focal buckets, they must be the same bucket. □

A partially ordered set can be represented by a DAG. We construct a graph with both directed and undirected edges, called an *order graph (OG)*, as follows: a node represents a bucket; for each pair of buckets $B$ and $B'$, there is a directed edge $B \longrightarrow B'$ if $B < B'$; there is an undirected edge $B\text{—}B'$ if the relation between them is "unknown". If we had a perfect statistical test for distributional changes, an OG would be a DAG. For the causal diagram shown in Figure 2(a) and the TS $(P, P_X, P_Y)$, the ideal OG is given in Figure 2(b).

In an OG, when $B$ is a focal bucket, a directed edge $B \longrightarrow B'$ asserts that there exists a directed path from the focal variable contained in $B$ to all the variables in $B'$. Hence, if there is no other *mixed directed path*, a path that could contain undirected edges but no directed edges in the reverse direction, from $B$ to $B'$ in the OG, there must be an edge from $B$ to at least one variable in $B'$ in the causal diagram. We mark this type of edges as $B \overset{*}{\longrightarrow} B'$, to distinguish them from those that only represent potential edges in the causal diagram. This information is useful when the child bucket $B'$ contains only one variable; we then assert that the edge $B \longrightarrow B'$ must exist in the causal diagram. We will call an OG with marked edges a *marked order graph* (MOG); an example is shown in Figure 2(c).

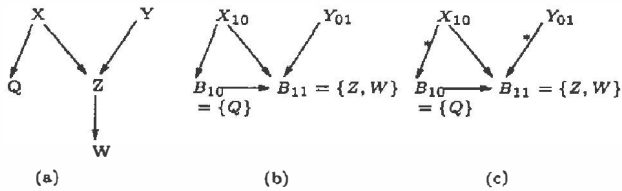

Figure 2: (a) A causal diagram; (b) The order graph for the TS $(P, P_X, P_Y)$; (c) The marked order graph.

An algorithm for constructing a MOG is given in the following.

**Algorithm 3 (Constructing MOG)**
*Input: an influential TS with known focal variables.*
*Output: a marked order graph.*

1. Put variables into buckets using Algorithm 1.

2. Extracting relations among buckets using Algorithm 2.

3. Let each bucket be a node.

4. For each pair of nodes $B$ and $B'$
   If $B < B'$, add an edge $B \longrightarrow B'$.
   If $B' < B$, add an edge $B' \longrightarrow B$.
   If the relation is "unknown", add an edge $B - B'$.

5. For each focal bucket $B^f$ and each of its child $B$
   If there is no other mixed directed path from $B^f$ to $B$, mark the edge as $B^f \stackrel{*}{\longrightarrow} B$.

In summary, the information conveyed by a MOG is as follows:

1. An unmarked edge $B \longrightarrow B'$: All variables in $B$ can be ordered before all variables in $B'$ in the causal diagram, in other words, there are no directed paths from variables in $B'$ to variables in $B$. When $B$ is a focal variable, there exists a directed path from $B$ to each variable in $B'$ in the causal diagram.

2. A marked edge $B \stackrel{*}{\longrightarrow} B'$: There exists a directed path from $B$ to each variable in $B'$. In the case that both $B$ and $B'$ contain one single variable, the edge $B \longrightarrow B'$ must exist in the causal diagram.

3. No edge between $B$ and $B'$: there is no directed path, in particular no edge, between any variable in $B$ and any variable in $B'$ in the causal diagram.

### 4.3 Limitation of detecting marginal changes

Can we fully recover a causal diagram by detecting marginal distribution changes alone? To fully recover a causal diagram, we must construct a MOG in which each bucket contains only one variable and every edge is marked. This may not, in general, be achieved. Considering a causal diagram $G$ containing a path $X \longrightarrow Z \longrightarrow Y$, it is clear that we can never determine if there is an edge $X \longrightarrow Y$ in $G$, since all marginal changes produced by transitions would be the same after adding that edge. What is the best we can get then by detecting marginal changes?

Given a DAG $G$, if we remove an edge $X \longrightarrow Y$ whenever there is a directed path from $X$ to $Y$, we get the *transitive reduction* of $G$. The transitive reduction of a DAG $G$ is the graph $G'$ with the fewest edges such that the *transitive closure* of $G'$ is equal to the transitive closure of $G$. The transitive closure of a DAG $G$ is the graph $G''$ such that an edge $X \longrightarrow Y$ is in $G''$ iff there is a directed path from $X$ to $Y$ in $G$. By detecting marginal changes in TS's, the best we can hope to get is the transitive reduction of the actual causal diagram. Since to mark an edge $X \longrightarrow Y$, $X$ must be a focal variable, it follows that every node except leaf nodes must be a focal variable in order to mark every edge in the transitive reduction graph. To further make each bucket contain only one variable, every leaf node having the same set of parents as another leaf node must be a focal variable.

In conclusion, by detecting marginal distribution changes, the best we can learn is the transitive reduction of the causal diagram, and we can achieve it by a TS in which every variable has had its mechanism changed.

### 4.4 Unknown focal variables

In this section we discuss situations where we know that a mechanism change has occurred at a single variable but we do not know the identity of that variable.

We first note that, without knowing the focal variables, variables can still be partitioned into buckets using Algorithm 1, and the relations between pairs of buckets will be determined by rules R1–R3 of Section 4.2. Second, an order graph can be constructed as follows: for each pair of buckets $B$ and $B'$, there is a directed edge $B \longrightarrow B'$ if $B < B'$. For the causal diagram of Figure 3(a) and the TS $(P, P_X, P_Y)$, the variables are partitioned into three buckets: $B_{10} = \{X, Q\}, B_{01} = \{Y\}, B_{11} = \{Z, W\}$, and the OG is shown in Figure 3(b).

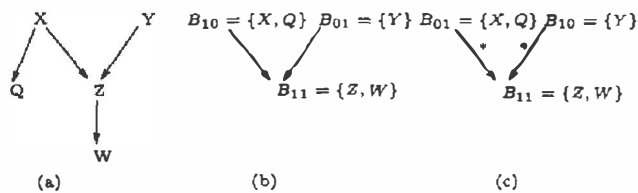

Figure 3: (a) A causal diagram; (b) The order graph for the TS $(P, P_X, P_Y)$ without knowing the focal variables; (c) The marked order graph.

Finally, we may be able to find to which bucket a focal variable belongs using the following theorem, assuming influentiality and perfect statistical tests. (We still call such a bucket a "focal bucket", because it behaves as a focal variable with the information at hand.)

**Theorem 6** *Let $S_j$ be the set of buckets for which $a_j = 1$ in their tags $a_1 \ldots a_k$, then the focal bucket $F^j$ for the $j$th transition is in $S_j$ and for any other bucket $B \in S_j$, $F^j < B$.*

**Proof:** Let the focal variable $X$ for the $j$th transition be tagged as $a_1 \ldots a_k$, then $a_j = 1$, since $P(X)$ must change in this transition. All other variables in the set of buckets $S_j$ must be descendants of $X$ since all their marginals changed in the $j$th transition. Therefore, whenever $P(X)$ changes, their marginals must change too, that is, if $a_i = 1$ then $b_i = 1$ for any variable tagged as $b_1 \ldots b_k$ in $S_j$, which leads to $a_i \leq b_i, i = 1, \ldots, k$. Hence for any bucket $B_{b_1 \ldots b_k} \in S_j$ not containing $X$, we have $B_{a_1 \ldots a_k} < B_{b_1 \ldots b_k}$. □

In practice, Theorem 6 may fail to identify a focal bucket when (due to imperfect statistical tests) there exists no bucket $F^j$ in $S_j$ satisfying $F^j < B$ for any other bucket $B \in S_j$. In the case that an identified focal bucket contains only one variable, we actually identify a focal variable. For the OG in Figure 3(b), the focal buckets for the first and second transitions can be found as $B_{10} = \{X, Q\}$ and $B_{01} = \{Y\}$ respectively, and we actually identify $Y$ as the focal variable of the second transition.

Finally we can get a MOG by marking edges as in Algorithm 3. For our working example, the ideal MOG is shown in Figure 3(c).

### 4.5 TSs absent of influentiality

If we allow for the possibility that a mechanism change at $X$ may not alter the marginal probabilities of some of $X$'s descendants, then detecting no change in $P(Y)$ provides no information on the causal relation between $X$ and $Y$. The information we may obtain is that detecting a change in $P(Y)$ means that $Y$ is a descendant of the focal variable $X$. First we partition variables into tagged buckets using Algorithm 1. Then the relationship among buckets is determined as: let $B^i$ be the focal bucket for the $i$th transition; $B^i < B_{a_1 \ldots a_k}$ if $a_i = 1$, where "<" represents that all variables in $B_{a_1 \ldots a_k}$ are descendants of the focal variable $B^i$. Finally we compute the transitive closure of $<$ relation, denoted by $<^*$, to get more information. Simultaneous $B <^* B'$ and $B' <^* B$ would mean change detection errors and the relation between $B$ and $B'$ will be declared as unknown. The information conveyed by $B <^* B'$ is that all variables in $B'$ are descendants of the focal variable $B$ in the underlying causal diagram.

It is clear that if the identities of the focal variables are not given, we can not get any order information from a TS by detecting marginal changes.

## 5 Combining Static and Dynamic Information

In Section 4, we discussed how to extract causal information given a TS by detecting distributional changes. In this section, we briefly describe how to combine this information with that obtained from independence tests.

Given data from a static stable distribution, we can recover (partially directed) causal diagrams using conditional independence tests. Several such algorithms have been developed, including IC algorithm [Pearl, 2000, section 2.5] (initially introduced in [Pearl and Verma, 1991]) and PC algorithm [Spirtes et al., 1993]. The output of these algorithms is a partially oriented graph representing an independence-equivalence class as defined by Theorem 1.

To recover a causal diagram from a TS, we first extract causal information by detecting distribution changes as described in Section 4, then run the IC algorithm using the causal information as prior knowledge. Note that since a TS is composed of a series of different distributions, we need to test independence relationships across all distributions.

We may obtain three types of causal information as shown in Section 4: causal order among certain variables, no edges between certain variables, and certain directed edges. The last two types (no-edge and determined-edge) can be incorporated directly. Causal order information can be used to restrict the search of candidate conditional sets and thus reduce the complexity of the IC algorithm. Causal order information can also be used to orient more edges: any undirected edge $X\text{---}Y$ can be oriented as $X \to Y$ if $X$ is ahead of $Y$ in the causal order. These methods of incorporating background knowledge have been discussed in [Spirtes et al., 1993, Section 5.4.5].

When the identities of all focal variables are known, after incorporating these causal information as background knowledge, the output of the IC Algorithm would be a partially oriented graph representing the TS equivalence class as defined by Theorem 3. This is due to a theorem in [Meek, 1995] which says that the orientation rules in the IC algorithm are complete with respect to any consistent background knowledge. If the identity of a focal variable is not given or identified as in Section 4.4, the edge directions between this focal variable and its neighbors may not be fixed, hence the output graph is not maximally oriented, and we have not obtained all the information implied by a TS. Algorithms for identifying focal variables are currently under investigation.

## 6 The Bayesian Approach

In the Bayesian approach, we compute the posterior probability of a causal diagram $G$ given a dataset $D$ as:

$$P(G|D,\xi) = \frac{P(D|G,\xi)P(G|\xi)}{P(D|\xi)}, \quad (8)$$

where $\xi$ represents our background knowledge. For the case that the dataset $D$ is from a static distribution, closed form expressions for $P(D|G,\xi)$ have been derived [Cooper and Herskovits, 1992, Heckerman et al., 1995]. In this section, we gave a closed form expression for $P(\mathbb{D}_{TS}|G,\xi)$. For detailed derivation, see [Tian and Pearl, 2001].

Let the sequence of datasets $\mathbb{D}_{TS} = \{D^0, D^1, \ldots, D^k\}$ be generated with parameters $\Theta_G^0, \ldots, \Theta_G^k$ respectively, and let $\Xi_G = \cup_{i=0}^k \Theta_G^i$. The marginal likelihood is computed as

$$P(\mathbb{D}_{TS}|G,F,\xi) = \int P(\mathbb{D}_{TS}|\Xi_G,G,\xi)P(\Xi_G|G,F,\xi)d\Xi_G. \quad (9)$$

We have put $F = (V_{i_1}, \ldots, V_{i_k})$ as a condition to reflect the fact that the sequence of focal variables are known. The term $P(\mathbb{D}_{TS}|\Xi_G,G,\xi)$ is computable as the probability of the data given a Bayesian network. For the parameter priors $P(\Xi_G|G,F,\xi)$, we use the assumptions given in [Heckerman et al., 1995]: *Global and Local Parameter Independence*, and *Parameter Modularity*, and we assume the following prior:

$$P(\Xi_G|G,F,\xi)$$
$$= P(\Theta_G^0|G,\xi)\Big(P(\Psi_{i_1}^1|G,\xi)\prod_{i\neq i_1}\delta(\Psi_i^1 - \Psi_i^0)\Big)$$
$$\Big(P(\Psi_{i_2}^2|G,\xi)\prod_{i\neq i_2}\delta(\Psi_i^2 - \Psi_i^1)\Big)$$
$$\cdots \Big(P(\Psi_{i_k}^k|G,\xi)\prod_{i\neq i_k}\delta(\Psi_i^k - \Psi_i^{k-1})\Big), \quad (10)$$

where $\delta(x)$ is the Dirac delta function, and we have used the notation $\Theta_G^j = \cup_{i=1}^n \Psi_i^j, j = 0, \ldots, k$. Eq. (10) says that the set of parameters $\Theta_G^j$ differs with $\Theta_G^{j-1}$ only by the parameters in $\Psi_{i_j}^j$, and we have made an assumption that the set of parameters $\Psi_{i_j}^j$ after a mechanism change is independent of the previous set of parameters $\Psi_{i_j}^{j-1}$. We assume the *Dirichlet distribution*:

$$P(\vec{\theta}_{pa_i}|\xi) = Dir(\vec{\theta}_{pa_i}|\vec{\alpha}_{pa_i}), \quad (11)$$

where $\vec{\alpha}_{pa_i} = \{\alpha_{v_i;pa_i}|v_i \in Dm(V_i)\}$ denotes the set of parameters for the Dirichlet distribution. Assuming that the set of parameters after a mechanism change have the same prior distribution as before: $P(\Psi_{i_j}^j|G,\xi) = P(\Psi_{i_j}^{j-1}|G,\xi)$, and that mechanism changes occurred at different variables, let $I = \{i_1, \ldots, i_k\}$ be the set of indexes for focal variables, and we obtain

$$P(\mathbb{D}_{TS}|G,F,\xi)$$
$$= \prod_{i\notin I}\prod_{pa_i} \frac{\Gamma(\alpha_{pa_i})}{\Gamma(\alpha_{pa_i} + M_{pa_i})} \prod_{v_i} \frac{\Gamma(\alpha_{v_i;pa_i} + M_{v_i,pa_i})}{\Gamma(\alpha_{v_i;pa_i})}$$
$$\times \prod_{l=1}^k \prod_{pa_{i_l}} \frac{\Gamma(\alpha_{pa_{i_l}})}{\Gamma(\alpha_{pa_{i_l}} + M_{pa_{i_l}}^l)} \prod_{v_{i_l}} \frac{\Gamma(\alpha_{v_{i_l};pa_{i_l}} + M_{v_{i_l},pa_{i_l}}^l)}{\Gamma(\alpha_{v_{i_l};pa_{i_l}})}$$
$$\times \prod_{l=1}^k \prod_{pa_{i_l}} \frac{\Gamma(\alpha_{pa_{i_l}})}{\Gamma(\alpha_{pa_{i_l}} + L_{pa_{i_l}}^l)} \prod_{v_{i_l}} \frac{\Gamma(\alpha_{v_{i_l};pa_{i_l}} + L_{v_{i_l},pa_{i_l}}^l)}{\Gamma(\alpha_{v_{i_l};pa_{i_l}})}, \quad (12)$$

where $\Gamma(\cdot)$ is the Gamma function, $\alpha_{pa_i} = \sum_{v_i} \alpha_{v_i;pa_i}$,

$$M_{v_i,pa_i}^l = \sum_{j=0}^{l-1} N_{v_i,pa_i}^j, \quad M_{v_i,pa_i} = M_{v_i,pa_i}^{k+1}, \quad L_{v_i,pa_i}^l = \sum_{j=l}^k N_{v_i,pa_i}^j,$$

$$M_{pa_i}^l = \sum_{v_i} M_{v_i,pa_i}^l, \quad M_{pa_i} = \sum_{v_i} M_{v_i,pa_i}, \quad L_{pa_i}^l = \sum_{v_i} L_{v_i,pa_i}^l,$$

and $N_{v_i,pa_i}^j$ is the number of cases in the dataset $D^j$ for which $V_i$ takes the value $v_i$ and its parents $Pa_i$ takes the value $pa_i$.

We will call the above Bayesian scoring metric $P(\mathbb{D}_{TS}, G|F,\xi)$ with parameters $\alpha_{v_i;pa_i}$ specified as required by the BDe metric in [Heckerman et al., 1995] the BDe_TS metric. A marginal likelihood $P(\mathbb{D}|G,\xi)$ is said to satisfy the property of $F$-transition likelihood equivalence if for two $F$-transition equivalent causal diagrams $G_1$ and $G_2$, $P(\mathbb{D}|G_1,\xi) = P(\mathbb{D}|G_2,\xi)$.

**Theorem 7** *The BDe_TS metric is $F$-transition likelihood equivalent.*

## 7 Experimental Results

We use $\chi^2$ test to detect distribution changes. Let $D^1$ and $D^2$ be two datasets, consisting of $N_1$ and $N_2$ cases respectively. Let $N_{1x}$ and $N_{2x}$ be the number of cases in $D^1$ and $D^2$ respectively in which a variable $X$ takes the value $x$. To test the hypothesis that $X$ has the same distribution in the two datasets, we compute the quantity

$$\chi^2 = N_1 N_2 \sum_x \frac{1}{N_{1x}+N_{2x}}(\frac{N_{1x}}{N_1} - \frac{N_{2x}}{N_2})^2, \quad (13)$$

which is asymptotically a $\chi^2$ distribution with $r_x - 1$ degree of freedom, where $r_x$ is the number of states of $X$. Let the significance level be $\alpha$. If $\chi^2 > \chi^2_\alpha$ then we decide "change", else we decide "no-change".

A mechanism change at a variable $V_i$ is simulated as follows. Consider parameters in $\vec{\theta}_{pa_i}$. If $\theta_{v_{i1};pa_i} \leq 0.5$ then let $\theta'_{v_{i1};pa_i} = \theta_{v_{i1};pa_i} + \delta$, else let $\theta'_{v_{i1};pa_i} = \theta_{v_{i1};pa_i} - \delta$, where $\delta$ is a parameter for adjusting the change magnitude. The rest of the parameters in $\vec{\theta}_{pa_i}$ are changed in proportional to their original values as: $\theta'_{v_{ij};pa_i} = \alpha \theta_{v_{ij};pa_i}, j = 2,\ldots,r_i$, where $\alpha = (1 - \theta'_{v_{i1};pa_i})/(1-\theta_{v_{i1};pa_i})$. When we simulate a mechanism change at $V_i$, we change parameters in $\vec{\theta}_{pa_i}$ as above for each $pa_i \in Dm(Pa_i)$.

In our experiments, we used data generated from a known network, the *Alarm* Bayesian network[1] [Beinlich et al., 1989]. Samples used in the experiment were generated from the network using a demo version of Netica API developed by Norsys Software Corporation. We used equal sample sizes for all datasets in a TS, that is, a sample size $N$ represents that $N$ cases were generated for each dataset $D^i$ in $\mathbb{D}_{TS} = \{D^0,\ldots,D^k\}$.

### 7.1 Errors in detecting changes

There are two types of errors in detecting changes: (i) mistaking "no-change" for a "change", known as type I error and denoted NC2C, and (ii) mistaking "change" as "no-change", known as type II error and denoted C2NC. Let $G$ be the causal diagram used for generating samples. When a mechanism change occurs at a variable $V_i$, if our test statistics is perfect, all $V_i$'s descendants in $G$ should be identified as "change" and $V_i$'s nondescendants as "no-change". Let $Dec_i$ be the number of descendants of $V_i$ in $G$ and $NDec_i$ be the number of nondescendants of $V_i$. Let $c2nc_i$ be the number of descendants of $V_i$ identified as "no-change" by the $\chi^2$ test, and let $nc2c_i$ be the number of nondescendants of $V_i$ identified as "change". $nc2c_i$ and $c2nc_i$ represent the number of type I and type II mistakes made by the $\chi^2$ statistics. In any one run, we simulate a mechanism change at each node $V_i, i = 1,\ldots,n$, relative to the *original* network, and compute the C2NC error rate as $\sum_i c2nc_i / \sum_i Dec_i$ and the NC2C error rate as $\sum_i nc2c_i / \sum_i NDec_i$. We computed an average error rate over 5 runs.

---
[1] We used the version downloaded from the web site of Norsys Software Corporation, http://www.norsys.com.

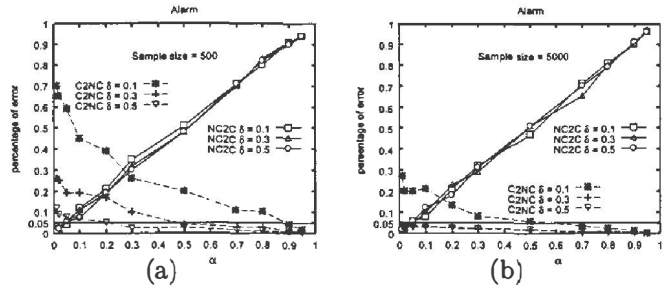

Figure 4: Type I and Type II errors of $\chi^2$ statistics.

We varied the change magnitude $\delta$, the sample size, and the significance level $\alpha$, and the results are shown in Figure 4. We see that the NC2C (type I) error rate is nearly the same as the $\alpha$ value for different change magnitudes and sample sizes, as expected. The C2NC (type II) error could be large when the $\alpha$ value is small or the change magnitude is small. This suggests that we should consider using a two-tailed $\chi^2$ test [Silverstein et al., 2000] to control the C2NC error, especially when the sample size is not large. In a two-tailed $\chi^2$ test, we use another threshold $\alpha' > \alpha$ such that we decide "no-change" only when $\chi^2 < \chi^2_{\alpha'}$, but we have to decide "unknown" when $\chi^2_{\alpha'} < \chi^2 < \chi^2_\alpha$. We will not discuss this method in this paper.

### 7.2 Errors in order graphs

In an OG, an edge $B \longrightarrow B'$ represents that all variables in $B$ can be causally ordered before the variables in $B'$. We call this type of information "order claims". No edge between $B$ and $B'$ represents the absence of directed paths, in particular edges, between variables in $B$ and those in $B'$; this information will be called "no-directed-path (NDP) claims" and "no-edge claims" respectively. An edge $B\!\!-\!\!B'$ only signals mistakes in the statistical tests and will be called "unknown claims". We performed the following experi-

<mark>UAI 2001</mark>

<mark>TIAN & PEARL</mark>

Table 1: Errors in order graphs. $k$: the number of focal variables. $m$: the number of buckets. $E_o$: percentage error of order claims. $E_p$: percentage error of NDP claims. $E_e$: percentage error of no-edge claims. $u$: number of unknown claims.

| | | | | $N = 500$ | | | | | |
|---|---|---|---|---|---|---|---|---|---|
| | | | | order claim | | NDP claim | | | |
| $k$ | $\delta$ | $\alpha$ | $m$ | # | $E_o$ | # | $E_p$ | $E_e$ | $u$ |
| 5 | 0.1 | 0.01 | 8 | 275 | 0.13 | 37 | 0.3 | 0.049 | 0 |
| 5 | 0.1 | 0.05 | 11 | 355 | 0.12 | 88 | 0.32 | 0.039 | 3 |
| 5 | 0.5 | 0.01 | 10 | 379 | 0.03 | 84 | 0.31 | 0.027 | 1 |
| 5 | 0.5 | 0.05 | 12 | 391 | 0.036 | 111 | 0.3 | 0.03 | 5 |
| 10 | 0.1 | 0.01 | 15 | 354 | 0.13 | 137 | 0.3 | 0.044 | 1 |
| 10 | 0.1 | 0.05 | 21 | 335 | 0.11 | 241 | 0.3 | 0.044 | 11 |
| 10 | 0.5 | 0.01 | 18 | 360 | 0.02 | 206 | 0.3 | 0.027 | 5 |
| 10 | 0.5 | 0.05 | 23 | 323 | 0.026 | 274 | 0.29 | 0.032 | 19 |
| | | | | $N = 5000$ | | | | | |
| | | | | order claim | | NDP claim | | | |
| $k$ | $\delta$ | $\alpha$ | $m$ | # | $E_o$ | # | $E_p$ | $E_e$ | $u$ |
| 5 | 0.1 | 0.01 | 10 | 369 | 0.044 | 80 | 0.3 | 0.025 | 2 |
| 5 | 0.1 | 0.05 | 12 | 393 | 0.051 | 109 | 0.3 | 0.031 | 5 |
| 5 | 0.5 | 0.01 | 10 | 400 | 0.014 | 78 | 0.19 | 0.015 | 2 |
| 5 | 0.5 | 0.05 | 12 | 406 | 0.026 | 104 | 0.26 | 0.027 | 7 |
| 10 | 0.1 | 0.01 | 19 | 364 | 0.027 | 207 | 0.28 | 0.02 | 6 |
| 10 | 0.1 | 0.05 | 23 | 334 | 0.029 | 260 | 0.28 | 0.033 | 20 |
| 10 | 0.5 | 0.01 | 19 | 377 | 0.0081 | 191 | 0.25 | 0.02 | 9 |
| 10 | 0.5 | 0.05 | 23 | 334 | 0.018 | 265 | 0.26 | 0.03 | 22 |

ments: for certain $\delta$, $\alpha$, sample size, and focal variables, we generate datasets, construct an OG, count the claims, and check against the true network to compute percentage errors for each type of claims.[2]

The results are shown in Table 1 for various sample size $N$, number of focal variables $k$, mechanism change magnitude $\delta$, and significance level $\alpha$. From Table 1, we see that the NDP claims have a high percentage of error; however, if those claims are interpreted as representing no-edge only, then the error rates are much lower. As expected, the error rates are lower when $\delta$, the change magnitude, is larger, and a TS with more focal variables produces more no-edge claims.

## 8 Conclusion

Spontaneous local changes offer the potential of extracting causal information that is undetected by static methods. This potential is limited by several factors, the most significant are violation of influentiality (in large networks) and the reliance on the locality of the changes. We believe that the former problem can be overcome by restricting the order information extracted to close neighborhoods of the focal variables.

### Acknowledgements

This research was supported in parts by grants from NSF, ONR and AFOSR and by a Microsoft Fellowship to the first author.

---

[2] Claims are counted between pairs of variables not between pairs of buckets. Numbers vary with the focal variables picked, hence we did 100 runs, each time randomly picking a sequence of $k$ variables as focal variables, and computed average numbers.